\documentclass[conference]{IEEEtran}
\IEEEoverridecommandlockouts
\usepackage{times}

\usepackage[numbers]{natbib}
\usepackage{multicol}
\usepackage[bookmarks=true]{hyperref}
\usepackage{amsfonts} 
\usepackage{xcolor}
\usepackage{graphicx}
\usepackage{amsmath}
\usepackage{booktabs} 
\usepackage{colortbl}
\usepackage{tabularray}
\usepackage{multirow}
\usepackage{fontawesome5}
\usepackage{tabularx}
\usepackage{array}
\usepackage{caption}
\usepackage{subcaption}
\usepackage{algorithm}

\DeclareMathOperator*{\argmin}{argmin}

\newcommand{\dmv}{\textsc{zlik}}

\newcommand{\anycar}{\textsc{AnyCar}}

\begin{document}

\title{\LARGE Zero-Shot Adaptation to Robot Structural Damage via \\ Natural Language-Informed Kinodynamics Modeling}


\author{\IEEEauthorblockN{
        Anuj Pokhrel$^1$, 
        Aniket Datar$^1$, 
        Mohammad Nazeri$^1$, 
        Francesco Cancelliere$^2$, and 
        Xuesu Xiao$^1$
    }
    \vspace{0.5em}
\IEEEauthorblockA{
        $^1$George Mason University\\
        $^2$ University of Catania\\
        \textcolor{magenta}{Code:} \href{https://github.com/AnujPokhrel/ZLIK}{\faGithub~GitHub} \quad \textcolor{magenta}{Video}: \href{https://www.youtube.com/watch?v=QbtaDONTQdQ}{\faYoutube~YouTube}
    }}

\makeatletter
 \g@addto@macro\@maketitle{
  \begin{figure}[H]
  \setlength{\linewidth}{\textwidth}
  \setlength{\hsize}{\textwidth}
  \centering
    \includegraphics[width=1\textwidth]{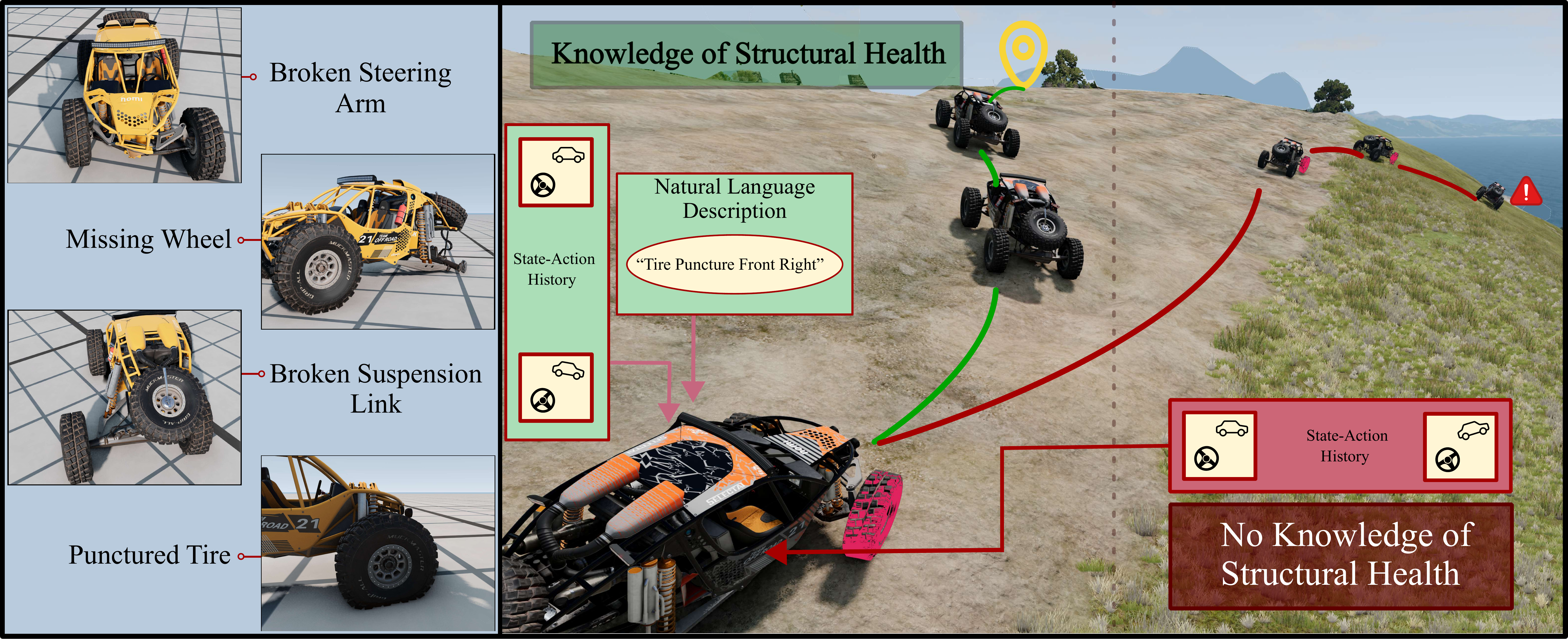}
    \caption{High-performance mobile robots may experience a variety of structural damages during in-the-wild operations (left), which significantly alter vehicle kinodynamics (right). When such kinodynamics discrepancies are not timely accounted for, catastrophic consequences may occur, e.g., falling off a cliff on the right due to a punctured front right tire (right). }
    \label{fig:motivation}
  \end{figure}
}
\makeatother
\maketitle
\addtocounter{figure}{-1}

\begin{abstract}
High-performance autonomous mobile robots endure significant mechanical stress during in-the-wild operations, e.g., driving at high speeds or over rugged terrain.
Although these platforms are engineered to withstand such conditions, mechanical degradation is inevitable.
Structural damage manifests as consistent and notable changes in kinodynamic behavior compared to a healthy vehicle.
Given the heterogeneous nature of structural failures, quantifying various damages to inform kinodynamics is challenging.
We posit that natural language can describe and thus capture this variety of damages.
Therefore, we propose Zero-shot Language Informed Kinodynamics (\dmv), which employs self-supervised learning to ground semantic information of damage descriptions in kinodynamic behaviors to learn a forward kinodynamics model in a data-driven manner.
Using the high-fidelity soft-body physics simulator BeamNG.tech, we collect data from a variety of structurally compromised vehicles.
Our learned model achieves zero-shot adaptation to different damages with up to 81\% reduction in kinodynamics error and generalizes across the sim-to-real and full-to-1/10$^{\text{th}}$ scale gaps.
\end{abstract}


\section{INTRODUCTION}
\label{sec::intro}

High-performance autonomous mobile robots deployed in the wild must contend with impact forces, torsional stress, and vibration caused by rapid and complex vehicle-terrain interactions. While modern platforms are engineered with robust chassis and suspension systems to mitigate these forces, the harsh nature of real-world environments makes mechanical degradation inevitable.
Consequently, robots frequently sustain structural damage ranging from punctured tires to bent steering columns and compromised suspensions (Fig.~\ref{fig:motivation} right). Nevertheless, mission-critical deployments frequently demand that robots tolerate and adapt to such damage to ensure timely task completion.

Such structural damage, however, drastically alters the vehicle’s kinodynamics by changing how forces are distributed and how the vehicle responds to control inputs. 
For example, a punctured tire isn’t able to generate as much traction as compared to other wheels, resulting in significant drag and rotational bias even when commanded with straight-line velocities. 
If the navigation system operates using a clean kinodynamics model while the physical platform is damaged, the discrepancy leads to tracking errors, instability, and potentially catastrophic control failure like collision with obstacles, vehicle rollover, or falling off a cliff (as shown in Fig.~\ref{fig:motivation} right). 


Accurately compensating for such kinodynamics mismatch poses a significant representation challenge.
Structural degradation manifests heterogeneously: it ranges from scalar parametric deviations (e.g., tire pressure loss) to complex geometric deformations (e.g., bent or broken suspension linkages) that fundamentally alter the robot's physical constraints.
Current state-of-the-art approaches to resolve the changing kinodynamics problem~\cite{xiao2025anycar} rely on adaptive control strategies or fine-tuning a previously learned kinodynamics model. However, these methods depend on an n-shot adaptation phase, requiring data collection through further vehicle-terrain interactions after the damage has occurred. This delayed process necessitates driving a structurally compromised robot with an incorrect kinodynamics model to collect training data, which is inherently risky and can worsen mechanical failure during data collection.

To bridge this gap without requiring dangerous online data collection and retraining, we propose a novel framework that leverages natural language to instantly adapt to vehicle structural damage in a zero-shot manner, i.e., Zero-shot Language Informed Kinodynamics (\dmv).  
We argue that, although the variety of structural damages cannot be represented by a single metric, they are semantically distinct and can be easily described using language.
Therefore, we incorporate vehicle damage into kinodynamics modeling through language descriptions. 
We utilize a Sentence Transformer to map natural language descriptions into a semantic embedding space, where damages exhibiting similar kinodynamic behaviors are clustered closer. These embeddings condition a kinodynamics model with spatiotemporal attention to predict the robot's future states based on the current robot health.
\dmv~allows the robot to instantly switch its internal kinodynamics model based on a diagnosis provided by an external vehicle health monitoring system, enabling zero-shot adaptation to different damages.
We validate \dmv~using a high-fidelity soft-body physics simulator, BeamNG.tech~\cite{beamng_tech} and deploy the system on a real-world robotic platform. Our contributions are as follows:

\begin{enumerate}
    \item A novel self-supervised learning approach that aligns language embeddings with kinodynamic responses in the presence of various vehicle structural damages;
    \item An accurate  kinodynamics model that utilizes the aligned language embeddings to represent heterogeneous structural damages for instant model adaptation; 
    \item Rigorous experiments including detailed studies on damage embeddings, model architecture, semantic grounding, and transfer from a simulated full-size vehicle to a physical 1/10$^{\text{th}}$ scale robot with different damages. 
\end{enumerate}

\section{RELATED WORK}
\label{sec::related}
We review related work in kinodynamics modeling, online adaptation, and robotic foundation models. 
\subsection{Kinodynamics Modeling}
Kinodynamics modeling has historically relied on analytical formulations~\cite{Pacejka01011992, rajamani2006vehicle}, such as bicycle or Ackermann steering models. 
These models offer computational efficiency but often oversimplify complex vehicle-terrain interactions~\cite{datar2023learning}. 
Therefore, researchers have predominantly shifted towards data-driven solutions~\cite{williams2018information, han2023model}.
To satisfy the high data requirements of these approaches without risking hardware, researchers frequently leverage high-fidelity simulators~\cite{todorov2012mujoco, beamng_tech, makoviychuk2021isaac, remonda2024simulation, tasora2016chrono}. 
To improve predictions beyond the training distributions~\cite{ovadia2019can}, recent works have incorporated uncertainty quantification using Gaussian Processes~\cite{ostafew2016robust, nagy2023ensemble, lee2023learning-based, ning2025dkmgp} or enforced kinematic constraints via Physics-Informed Neural Networks~\cite{raissi2019physics, maheshwari2023piaug, zhao2024physord, cai2025pietra}. 
Regardless of the model architecture, accurate prediction relies on information provided by the input modalities.

Input modalities in existing works have primarily focused on visual semantics~\cite{ostafew2016robust, wigness2019rugd}, elevation mapping~\cite{datar2023learning, datar2024terrain, nazeri2025verticoder, nazeri2025vertiformer}, terrain geometry~\cite{meng2023terrainnet, gibson2025dynamics}, and inertial measurements~\cite{xiao2021learning, karnan2022vi, pokhrel2024cahsor} to perceive the external environment and inform kinodynamics. 
However, changes in robot's internal physical state such as structural damages also play a critical role in determining kinodynamics. 
Structural damages can take a variety of forms, which can or cannot be measured by commonly used sensor suites and manifest by both quantitative measurements and qualitative descriptions. Therefore, we posit natural language descriptions provide the necessary semantic signal to represent a robot's internal health conditions and can cover a variety of heterogeneous vehicle damages that also affect kinodynamics. 

\subsection{Online Adaptation}
Traditional online system identification methods are designed to handle parametric variations in complex environments, such as estimating friction coefficients~\cite{rabiee2019friction, nagy2023ensemble} or slip ratios~\cite{yu2023fully}. 
However, these approaches assume a fixed vehicle structure and treat deviations as scalar parameter changes within a known physics model. 
In contrast, structural damages or mechanical failures induce large kinodynamics shifts that cannot be captured by simple parameter variations~\cite{cully2015robots}. 

To address non-linear dynamics shifts, researchers have relied on meta-learning and online adaptation methods~\cite{wang2024pay}. 
Approaches such as Model-Agnostic Meta Learning~\cite{finn2017model}, Meta-Reinforcement Learning~\cite{rakelly2019efficient}, Function Encoders~\cite{ingebrand2025function}, and CaRoL~\cite{11197900} treat dynamics shifts as new tasks, updating the base model’s weights through gradient-based updates using recent state-action pairs. 
These approaches require new physical interactions to infer new dynamics and introduce adaptation delays in an n-shot fashion~\cite{nagabandi2018learning, lupu2024magic, levy2025meta}. 
Such continuous environment interactions using an incorrect kinodynamics model caused by delayed adaptation can lead to unsafe maneuvers on a mechanically compromised platform. 

In contrast, \dmv~entirely abandons the paradigm of online retraining with new data. Instead, we frame the problem as conditional modeling. By training our model on a diverse manifold of damage descriptions, \dmv~learns a fixed set of weights that can generalize across different structural health states. During deployment, the dynamics model's weights do not need to be updated through dangerous data collection and delayed weight adaptation, rather it simply switches its predictive context based on the natural language descriptions of the damage. This zero-shot adaptability allows the robot to respect its new physical limits the moment a damage description is provided.

\subsection{Robotic Foundation Models}

Using pre-trained models to accelerate learning is a well-established concept in robotics. Data-driven approaches have heavily relied on visual backbones pre-trained on large-scale datasets to extract robust environmental features~\cite{jung2024v}. 
Architectures such as ResNet~\cite{He2016} and SegFormer~\cite{xie2021segformer} have become standard encoders for tasks ranging from semantic segmentation~\cite{shaban2022semantic} to traversability estimation~\cite{triest2024velociraptor}. 

The introduction of the Transformer architecture~\cite{vaswani2017attention} has shifted the paradigm from simple feature extraction to holistic foundation models~\cite{kirillov2023segment, yang2024depth}. 
In the context of dynamics modeling, this architecture is especially efficient in capturing temporal context and long-horizon dependencies~\cite{lotfi2024uncertainty, nazeri2025verticoder, xiao2025anycar}.
However, standard robotic implementations typically tokenize the entire state tuple (x, y, z, roll, pitch, yaw) as a single input embedding~\cite{nazeri2025vertiformer}.
This monolithic approach obscures the independence of state dimensions.
To better capture the nuances of structural damages where specific degrees of freedom may be affected disproportionately, we decompose the state vector into dimension-specific tokens, allowing the attention mechanism to isolate compromised dynamic channels~\cite{grigsby2021long, zhang2023crossformer}.

While Transformers have facilitated the use of Vision-Language Models~\cite{radford2021learning, liu2023visual} and Vision-Language Action Models~\cite{brohan2023rt, shah2023lm, kawaharazuka2025vision}, the use of language in these frameworks is primarily for high-level reasoning~\cite{min2025advancing, elnoor2025vlm, zhang2026vision}. 
To the best of our knowledge, language hasn't been utilized to describe the intrinsic physical state of the robot itself to inform low-level kinodynamics modeling. Our work bridges this gap by leveraging the semantic richness of pre-trained language models to encode physical variations into robot kinodynamics.

\section{APPROACH}
\label{sec::approach}

Our objective is to learn a kinodynamics model that accurately predicts the future state of a mechanically compromised robot. We formulate this as learning a kinodynamics function conditioned on the robot's state-action history and a textual description of the damage. 

\subsection{Problem Formulation}

We define the robot's state at time $t$ as $\mathbf{s}_t \in \mathcal{S} \subset \mathbb{SE}(3)$, representing the 6-Degree of Freedom (DoF) pose: $\mathbf{s}_t = [x_t, y_t, z_t, \textrm{roll}_t, \textrm{pitch}_t,\textrm{yaw}_t]^\top$. The control input applied to the robot, i.e., linear and angular velocities, is denoted by $\mathbf{u}_t \in \mathcal{U}$: $\mathbf{u}_t = [v_t, \omega_t]^\top$.
Under nominal conditions, the evolution of the robot's state on uniform terrain is governed by a nominal forward kinodynamics function $f_\textrm{nominal}: \mathcal{S} \times \mathcal{U} \rightarrow \mathcal{S}$:
\begin{equation}
\mathbf{s}_{t+1} = f_\textrm{nominal}(\mathbf{s}_t, \mathbf{u}_t).
\end{equation}
Physically, this function encapsulates how the robot's chassis, suspension, and powertrain distribute forces.
The chassis distributes external impact forces across a mesh of beams and nodes, while the suspension dampens vertical impulses to maintain consistent tire contact. 
In a healthy state, the mapping from $\mathbf{u}_t$ to the resultant forces acting on the center of mass remains unbiased and follows standard rigid body mechanics. 

However, when a robot incurs structural damage, this physical mapping is fundamentally altered.
A specific damage instance, denoted as $d \in \mathcal{D}$, breaks the nominal force distribution pathways in the chassis or energy transfer pathways in the powertrain.  
For example, a snapped half-shaft prevents the powertrain from transferring torque to a specific wheel, while a compromised suspension linkage alters how ground reaction forces are resolved into the chassis. These mechanical failures introduce a non-linear disturbance to the system dynamics.

\begin{figure*}[t]
    \centering
    \begin{subfigure}[t]{0.38\textwidth}
        \centering
        \includegraphics[height=0.116\textheight]{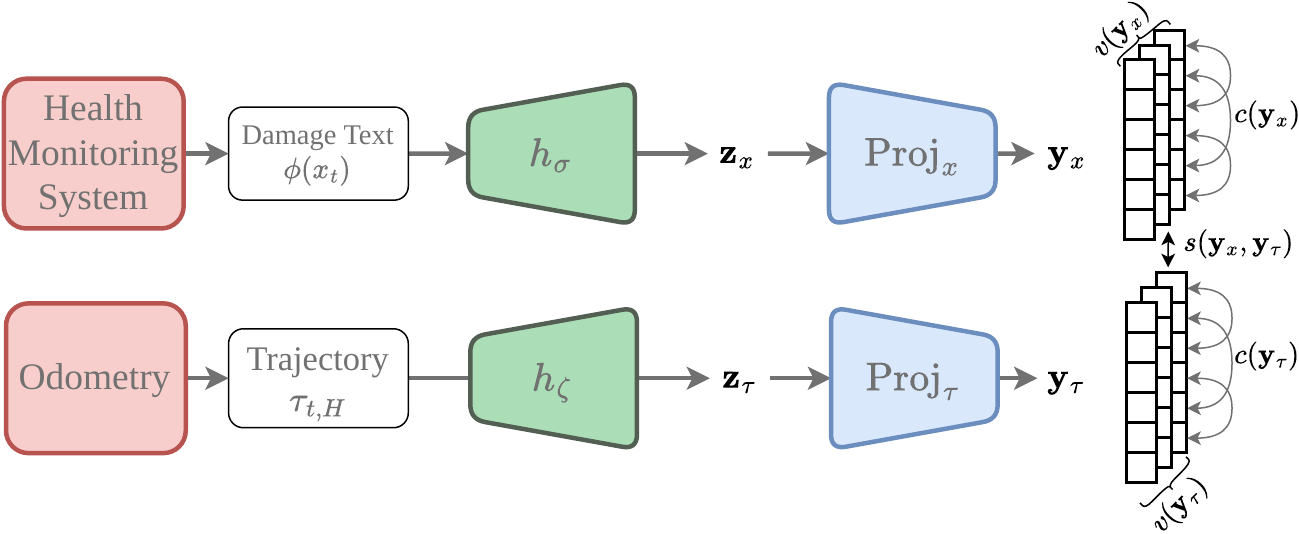}
        \caption{Damage Representation.}\label{fig:damage_representation}
    \end{subfigure}%
    \begin{subfigure}[t]{0.62\textwidth}
        \centering
        \includegraphics[height=0.116\textheight]{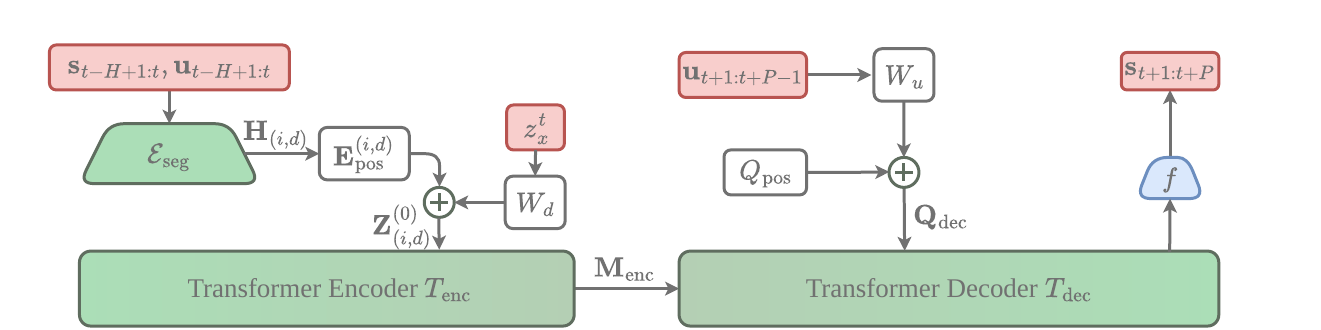}
        \caption{\dmv~Model Architecture.}\label{fig:architecture}
    \end{subfigure}
    \caption{(a) Constructing a representation space for damage that is both semantically descriptive and kinodynamically grounded. (b) \dmv~leverages a Transformer Encoder-Decoder structure to approximate damaged kinodynamics. }\label{fig:training}
\end{figure*}

We posit that while these disturbances are complex, they manifest as consistent, distinct patterns in robot's kinodynamic behavior. For example, a burst front left tire will cause vertical impulses in the front left side of the vehicle along with lateral movement towards the left. 
To account for these deviations, we extend the kinodynamics formulation to depend explicitly on the damage instance $d_t$ at time $t$.
Furthermore, because the dynamic effects of damage often exhibit temporal dependencies that cannot be easily captured in a single state transition, we condition our model on a history length $H$.
We formulate the damage-aware forward kinodynamics as a function $f_\textrm{damage-aware}: \mathcal{S}^{H} \times \mathcal{U}^H \times \mathcal{D} \rightarrow S$:
\begin{equation}
\mathbf{s}_{t+1} = f_\textrm{damage-aware}(\mathbf{s}_{t-H+1:t}, \mathbf{u}_{t-H+1:t}, d_t).
\label{eqn::damaged_fkd}
\end{equation}
Here, $d_t$ acts as a latent variable that effectively switches the context of the prediction based on the current damage.
The core challenge then becomes how to define and use $d_t$ for kinodynamic predictions.
In our approach, we aim to construct a representation space for damage that is both semantically descriptive and kinodynamically grounded (see Fig.~\ref{fig:damage_representation}). 

\subsection{Damage Representation}

To enable zero-shot adaptation, our damage representation space aims to bridge the gap between natural language descriptions and kinodynamic behaviors.
Since characterizing heterogeneous structural damages via a single scalar metric is difficult, we use a natural language description of the robot's health state at time $t$, denoted as $x_{t}$.
We employ a pre-trained sentence Transformer $\phi$ to encode this $x_t$ into a semantic embedding.
Sentence Transformers are optimized such that semantically similar damage descriptions result in spatially proximal embedding vectors, preserving the linguistic meaning of the damage description.

However, linguistic proximity does not always guarantee kinodynamic proximity. For example, a ``broken axle" and a ``snapped half-shaft" are linguistically different but kinodynamically identical.
To bridge this gap, we must align the linguistic semantic space with the physical behavioral space.
We posit that damages manifest as consistent, distinct patterns in the robot's state-action trajectory. 
We define the robot's kinodynamic behavior at time $t$ as its state-action trajectory over the history length $H$:
\begin{equation}
\tau_{t, H} = ((s_{t-H+1}, u_{t-H+1}),....,(s_t, u_t))\in \mathbb{R}^{(6+2) \times H}.
\nonumber
\end{equation}
Our objective is to construct a shared representation space $\mathcal{Z}$ where the semantic encoding of the damage is structurally aligned with its induced kinodynamic behavior. 
We use Multi-Layer Perceptrons (MLP) to map the natural language description of the health state $x$ and kinodynamic behavior $\tau$ into this shared space, resulting in $\mathbf{z}_x = h_\sigma(\phi(x))$ and $\mathbf{z}_{\tau} = h_{\zeta}(\tau)$.

As shown in Fig.~\ref{fig:damage_representation}, we align the projection of these representations, denoted as $\mathbf{y}$, using Variance-Invariance-Covariance Regularization (VICReg) ~\cite{bardes2021vicreg}. This loss function forces the projected text embedding $\mathbf{z}_x$ to be predictive of the kinodynamic behavior $\mathbf{z}_{\tau}$.
The loss is defined as a weighted sum of three terms:
\begin{equation}
\scriptstyle
\mathcal{L}_\textrm{align}(\mathbf{y}_{x}, \mathbf{y}_{\tau}) = \overbrace{\lambda s(\mathbf{y}_{x}, \mathbf{y}_{\tau})}^{\text{Invariance}} + \overbrace{\mu [v(\mathbf{y}_{x}) + v(\mathbf{y}_{\tau})]}^{\text{Variance}} + \overbrace{\nu [c(\mathbf{y}_{x}) + c(\mathbf{y}_{\tau})]}^{\text{Covariance}}, \nonumber
\end{equation}
where $\lambda$, $\mu$, and $\nu$ are hyperparameters controlling the importance of each term.
Here, the invariance term minimizes the Mean Squared Error (MSE) between the semantic damage description and the observed kinodynamic behavior.
This forces the language embedding to be a proxy for the physical reality of the damage:
\[s(\mathbf{y}_{x}, \mathbf{y}_{\tau}) = \frac{1}{N} \sum_{i=1}^{N} \| \mathbf{y}_{x}^{(i)} - \mathbf{y}_{\tau}^{(i)} \|_2^2,\]
where $N$ is the batch size.
Furthermore, the variance term ensures that the damage embeddings maintain diversity across the batch, capturing heterogeneity of structural failures. 
This prevents the model from collapsing to a trivial solution where all the embeddings map to a single point with zero variance.
Variance of each embedding dimension is forced to be above a threshold $\gamma$ by a hinge loss on the standard deviation $\sigma$:
\[ v(\mathbf{y}) = \frac{1}{K} \sum_{j=1}^{K} \max(0, \gamma - \sigma(\mathbf{y}^j, \epsilon)), \]
where $\sigma(x, \epsilon) = \sqrt{\text{Var}(x) + \epsilon}$, $\epsilon$ is a small scalar value to prevent numerical instabilities, and $K$ is the total number of dimensions in the embedding vector $\mathbf{y}$.
Finally, the covariance term ensures the damage embedding captures independent factors of the dynamics rather than redundant correlations. 
We define the covariance matrix $C(\mathbf{y})$ over the batch dimension $N$ to capture the correlation between embedding features:
\[C(\mathbf{y}) = \frac{1}{N-1} \sum_{i=1}^{N} (\mathbf{y}^{(i)} - \bar{\mathbf{y}})(\mathbf{y}^{(i)} - \bar{\mathbf{y}})^T \textrm{, where} ~\bar{\mathbf{y}} = \frac{1}{n} \sum_{i=1}^{n} \mathbf{y}^i \textrm{.}\]
The sum of squared off-diagonal coefficients are minimized to de-correlate the dimensions: 
\[ c(\mathbf{y}) = \frac{1}{K} \sum_{i \neq j} [C(\mathbf{y})]_{i,j}^2. \]
By forcing the off-diagonal coefficients to zero, we maximize the information content of the damage representation, allowing the downstream kinodynamics model to leverage independent axes of variation.
The scale factor of $1/K$ ensures the covariance criterion scales linearly with the dimensionality.


Finally, given a dataset of $N$ paired samples $\{(x_{i}, \tau_{i})\}_{i=1}^{N}$, we seek the optimal encoder parameters $\sigma^*$ and $\zeta^*$ to minimize this loss expectation:
\begin{equation}
\sigma^*, \zeta^* = \argmin_{\sigma, \zeta} \frac{1}{N} \sum_{i=1}^{N} \mathcal{L}_\textrm{align}(h_\sigma(\phi(x_i)), h_{\zeta}(\tau_i)).
\label{eqn:minvicreg}
\end{equation}
This formulation ensures that the learned embedding space $\mathcal{Z}$ is robust across diverse structural damages. 

\subsection{Downstream Kinodynamics Modeling}

After the self-supervised training to correlate the damage representation and kinodynamic behavior, we freeze the projection head $h_{\sigma}$. We replace the damage instance $d_t$ in Eqn. \eqref{eqn::damaged_fkd} with encoded representation of the damage instance's language description, $\mathbf{z}_x^t$.
We assume language description of damage $x_t$ is provided by a vehicle health monitoring system.
By utilizing the frozen projection head $h_{\sigma}$, we learn to approximate forward kinodynamic function $f_{\theta}(\cdot)$ as a downstream task.
The forward kinodynamic function is given by:
\begin{equation}
\begin{split}
    & \underbrace{\mathbf{s}_{t+1:t+P}}_{\text{Predicted Future States}} =  f_{\theta}(\underbrace{\mathbf{s}_{t-H+1:t}, \mathbf{u}_{t-H+1:t}}_{\text{History}}, \underbrace{\mathbf{u}_{t+1:t+P-1}}_{\text{Future Actions}}, \mathbf{z}_x^t),
    \label{eqn::dmv}
\end{split}
\end{equation}
where $\theta$ is learnable parameters and $P$ is the prediction horizon. We learn the optimal parameters $\theta^{*}$ in a supervised fashion using our dataset of damaged robot operation in BeamNG.tech, $\{\mathbf{s}^j_{t+1:t+P}, \mathbf{s}^j_{t-H+1:t}, \mathbf{u}^j_{t-H+1:t},
\mathbf{u}^j_{t+1:t+P-1},
\mathbf{z}_x^{t, j}\}_{j=1}^{N}$.
\vspace{3pt}
We seek the optimal kinodynamics parameters $\theta^*$:


\begin{equation}
\begin{split}
    \theta^{*} = \argmin_{\theta} \sum_{j=1}^{N} \Bigg\| & \overbrace{\mathbf{s}^{j}_{t+1:t+P}}^{\text{Actual Future States}} - \\
    & f_{\theta}(\underbrace{\mathbf{s}^{j}_{t-H+1:t}, \mathbf{u}^{j}_{t-H+1:t}}_{\text{History}}, \underbrace{\mathbf{u}^{j}_{t+1:t+P-1}}_{\text{Future Actions}}, \mathbf{z}_x^{t,j}) \Bigg\|^2.
\end{split}
\label{fkd_loss}
\nonumber
\end{equation}

\subsection{Spatiotemporal Attention}

\begin{figure}[t]
    \includegraphics[width=\columnwidth]{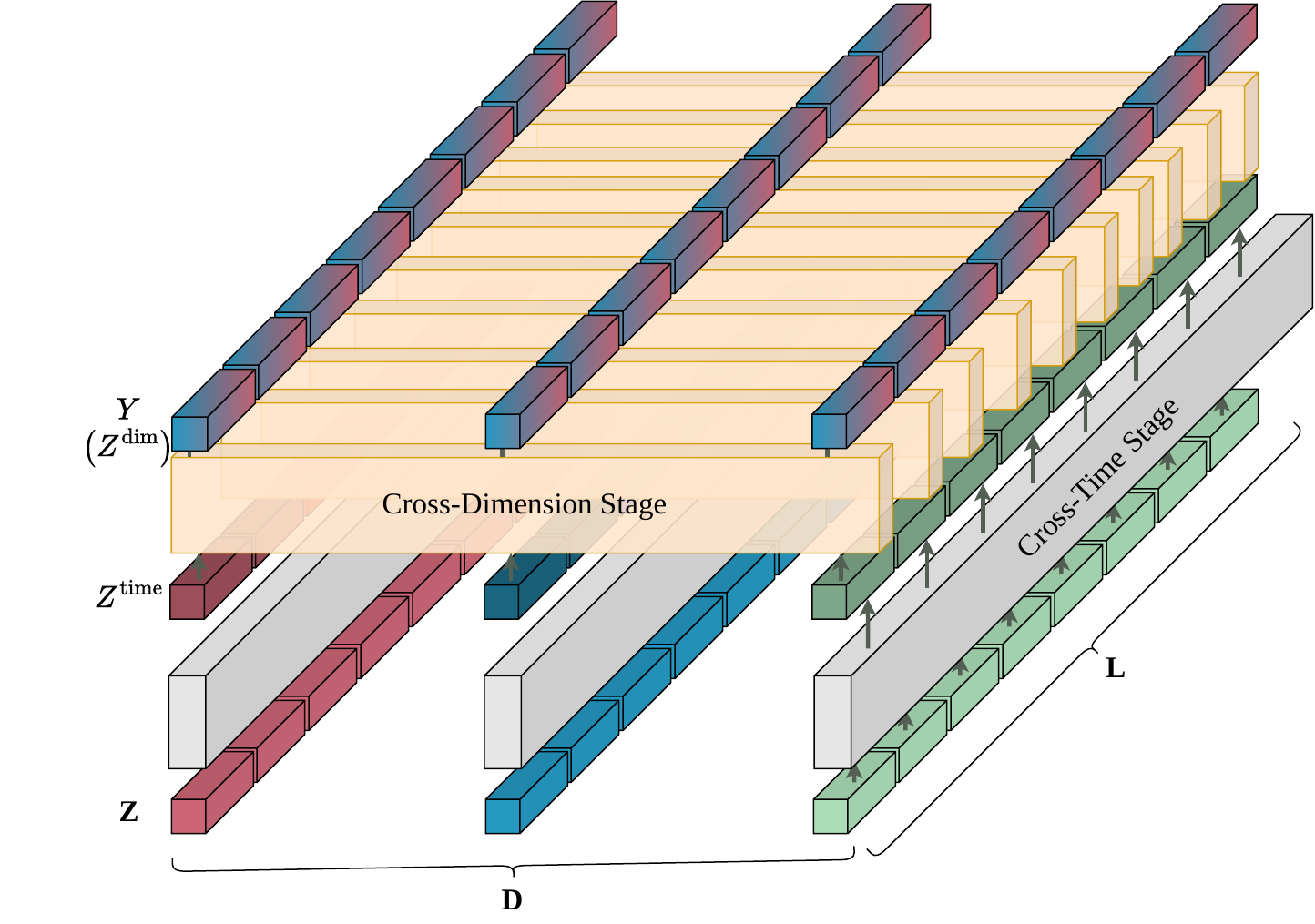}
    \caption{\dmv's Transformer Encoder uses a two-stage attention architecture for both time (gray) and state dimension (orange).}
    \label{fig:two_stage_attention}
\end{figure}

We approximate the kinodynamic function in Eqn. \eqref{eqn::dmv} using a Transformer Encoder-Decoder architecture (Fig.~\ref{fig:architecture}), similar to Crossformer~\cite{zhang2023crossformer}, designed to capture spatiotemporal dependencies as shown in Fig.~\ref{fig:two_stage_attention}. 
To tokenize the robot's state history for this model, we employ a Dimension-Segment Embedding, denoted as $\mathcal{E}_{\text{seg}}$.
This embedding transforms the raw state-action history $\tau_{t, H}$ into a structured feature array $\textbf{H}$ by preserving the independence of state dimensions.

Specifically, $\mathcal{E}_{\text{seg}}$ partitions the temporal history of each dimension into local windows of length $L_{\text{seg}}$, which are then projected into the Transformer's latent dimension $d_{\text{model}}$. 
This results in a 2D feature array $\mathbf{H} \in \mathbb{R}^{L \times D \times d_{\text{model}}}$, where $D$ represents the state dimensions and $L=H/L_{\text{seg}}$ denotes the number of resulting time segments (Fig.~\ref{fig:two_stage_attention} bottom).

To condition the dynamics on the robot's structural health, as shown in Fig.~\ref{fig:architecture}, we inject the semantic damage embedding $\mathbf{z}_x \in \mathbb{R}^{d_{\text{damage}}}$ directly into this latent space along with learnable positional embedding to create the input to the encoder layer, $$\mathbf{Z}_{i,d}^{(0)} = \mathbf{H}_{i,d} + \mathbf{E}_\textrm{pos}^{(i,d)} + W_{\text{d}} \mathbf{z}_x.$$
The indices $(i, d)$ correspond to the time segment and state dimension. $\mathbf{E}_\textrm{pos} \in \mathbb{R}^{d_{\text{model}}}$ is a learnable positional embedding, and $W_{\text{d}} \in \mathbb{R}^{d_{\text{model}} \times d_{\text{damage}}}$ is a learnable projection matrix. By adding the damage context $W_{\text{d}} \mathbf{z}_x$ globally to every segment $(i, d)$ we bias the robot’s entire history with semantic signature of the damage.

The Transformer Encoder $T_\textrm{enc}$ processes $\mathbf{Z}^{(0)}$ through a stack of two-stage attention layers to capture the dependencies along  the spatial and temporal dimensions.
First, the model uses Multi-head Self-Attention (MSA) across time segments $i$ to capture the temporal evolution of each dimension $d$ (Fig.~\ref{fig:two_stage_attention} gray).
Second, the model captures inter-variate dependencies by applying MSA across dimensions $d$ for each time segment $i$ (Fig.~\ref{fig:two_stage_attention} orange).
This disentangled attention mechanism enables the network to learn how the injected damage $\mathbf{z}_x$ specifically alters the correlations between physical dimensions. We denote the output of the encoder as the damage-informed memory, $\mathbf{M}_\textrm{enc} = T_\textrm{enc}(\mathbf{Z}^{(0)})$.

To predict the future trajectory, we employ a non-auto-regressive Transformer decoder $T_\textrm{dec}$ conditioned on the sequence of future actions $\mathbf{u}_{t+1:t+P-1}$.
We project the future actions into a latent space using a learnable projection matrix $W_{u}$ and then add to a learnable positional embedding $\mathbf{Q}_\textrm{pos}$, generating the decoder query
$\mathbf{Q}_\textrm{dec} = \mathbf{Q}_\textrm{pos} + W_u \mathbf{u}_{t+1:t+P-1}$.
The decoder performs cross-attention on the given query to retrieve relevant dynamics from the encoder's damage-informed memory $\mathbf{M}_\textrm{enc}$. 
The final decoder output is projected to the state dimension via a linear head $\hat{\mathbf{s}}_{t+1:t+P} = f(T_\textrm{dec}(\mathbf{Q}_{\text{dec}}, \mathbf{M_{\text{enc}}})) \in \mathbb{R}^{P \times 6}$.
\section{IMPLEMENTATIONS}
\label{sec::implementations}
In this section, we present implementation details of our approach and experiments. 

\subsection{Simulation, Dataset, Robot, and Damages}
For data generation and experimentation, we use a high-fidelity soft-body physics simulator, BeamNG.tech, which has been extensively used for dynamics simulation in off-road environments.
The vehicles in BeamNG.tech are built as a network of nodes and meshes, while also simulating the elasticity and strength of each mechanical part. 
In our experiment, we use a full-size vehicle as our robot.
The simulator maintains an internal robot health state accessible in JSON format. We convert the relevant JSON keys and values into natural language descriptions $x_t$. 

Our dataset includes six damage classes including single and a combination of broken part(s):
\begin{itemize}
    \item \textbf{Tire Puncture}: a single punctured tire;
    \item \textbf{Tire \& Spring}: a single punctured tire along with a broken suspension on the same wheel;
    \item \textbf{Multiple Tires Punctured \& Suspensions Broken (MTP \& SB)}: damages to adjacent tires and suspensions;
    \item \textbf{Broken Axle}: either of the axles broken;
    \item \textbf{Fall}: Drop the robot from a random height in $[5, 15]$ meters at a random orientation to incur multiple damages;
    \item \textbf{No Damage}: healthy vehicle. 
\end{itemize}

The mechanically compromised robots are then operated using a random walk algorithm. We collect vehicle state and action to generate training data for both damage representation pre-training and downstream kinodynamics learning.   
The dataset is collected uniformly across all classes.
In total, we collect 150K data points, which are divided into an 80-20 train-validation split. Another unseen dataset of 30K is collected for testing, which are used to report all experiment results. 

For real-world experiments, we use an open-source, 1/10$^{\text{th}}$ scale Verti-4-Wheeler platform, V4W~\cite{datar2024toward}.
Real-time state information is provided at 100Hz by a motion capture setup. 
The V4W is damaged by manually removing a wheel or the foam material inside a tire.
All kinodynamics training and inference is conducted on an NVIDIA A6000 GPU with 48GB of RAM wirelessly connected with the robot.


\subsection{Damage Representation}

To encode the natural language damage descriptions, we utilize the EmbeddingGemma (300M)~\cite{vera2025embeddinggemma} model as our sentence Transformer backbone $\phi$.
The model is chosen for its balance of lightweight inference and good semantic understanding. 

The kinodynamic trajectory encoder $h_\zeta$ is implemented as a Transformer Encoder. 
The encoder processes a history window of $H=200$ steps (10 seconds), where the input at each step is the concatenation of the relative pose in the robot frame and executed action. 
We utilize a learnable positional encoding to preserve temporal order.
The encoder consists of 3 Transformer layers with 8 attention heads, a hidden dimension of 128, and a feed-forward dimension of 512. We apply a dropout of 0.2 for regularization. 
The output sequence is aggregated via global average pooling to extract a fixed-size feature vector.


Both the text embedding and the trajectory encoding are projected into a shared latent space of dimension K = 128 using three-layer MLPs with 256, 256, 128 hidden dimensions, Batch Normalization, and ReLU activations.

 We align the semantic  and kinodynamic representations, $\textbf{z}_x~\text{and}~ \mathbf{z}_\tau$, using Eqn~\eqref{eqn:minvicreg}. The hyperparameters for the loss function are set to $\lambda = 25.0, \mu = 10.0\text{, and } \nu = 0.1$ after hyperparameter tuning.

 \subsection{Damaged Kinodynamics Model}

Our Transformer implementation takes as input the state-action pairs over a history $H = 40$ (2 seconds), a sequence of future actions of length $P=10$ (0.5 seconds), and the semantic damage embedding $\mathbf{z}_{x}^{t}$ as the input.

The input state history is segmented into patches of length $L_\textrm{seg} = 4$, resulting in 10 segments per state dimension.
These segments are projected into a latent dimension of $d_{\text{model}} = 256$. The learnable positional embedding added to these segments preserves both temporal ordering and state dimension identity.
The 128-dimensional damage embedding $\mathbf{z}_x$ is projected via a linear layer to match $d_\text{model}$ and is added element-wise to the position-encoded segments to condition the global context.
The Transformer Encoder consists of $N = 3$ layers utilizing two-stage attention.
We implement the temporal attention phase using a local window size of $W_\textrm{size} = 2$ to efficiently capture local dependencies. 
The Transformer Decoder consists of $N=4$ layers.
To condition the generation on control inputs, the future action sequence $\mathbf{u}_{t+1:t+P-1}$ is segmented, projected to $d_{\text{model}}$ via a linear layer, and added directly to the Decoder's learnable positional query embeddings. 
The Decoder then attends to the Encoder memory via cross-attention.

The model is implemented in PyTorch. We set $h=4$ attention heads, a routing factor of $c=10$, and a feed-forward dimension of $d_\textrm{ff} = 512$ for both Encoder and Decoder layers.
We apply a dropout of 0.2 throughout the network.
The final Decoder output is projected to the state dimension via a linear head, and the network is trained to minimize MSE between the ground truth and predicted trajectories in the next 0.5 seconds.

\section{EXPERIMENTS}

We design our experiments to answer four core research questions regarding the efficacy of \dmv.

\subsection{Research Question 1: Zero-Shot Performance}
In this experiment, we demonstrate \dmv's ability to achieve zero-shot adaptation to structural failures and validate the necessity of explicit damage representation in kinodynamics modeling for structurally compromised robots. We compare \dmv's trajectory prediction performance against a non-adaptive and a state-of-the-art adaptive baseline, \anycar~\cite{xiao2025anycar}.
The non-adaptive model employs a Transformer Encoder-Decoder architecture similar to our backbone but is trained exclusively on state-action data from a structurally healthy robot. 
This model does not receive damage embedding as input, thus named CleanTransformer. 
We utilize the \anycar~model~\cite{xiao2025anycar} as a baseline for kinodynamics adaptability.
Like the CleanTransformer, the \anycar~model does not utilize damage embedding and is also trained on state-action data of the healthy robot.

To evaluate the efficacy of online adaptation versus our zero-shot approach, we fine-tune the \anycar~model on new data collected on damaged robots. 
For a fair comparison with \dmv's adaptability, we fine-tune \anycar~with only 20 seconds of data. To show \anycar's full capability with larger amount of new iteration data collected after damage has occurred, we also fine-tune it on 5 and 10 minutes of new data.


\begin{table}[h]
\centering
\renewcommand{\arraystretch}{1.3}
\caption{Comparison of Trajectory Prediction Error across All Damage Classes. \dmv~achieves the lowest error in zero-shot in every scenario, even outperforming models with 10 minutes of environment-specific fine-tuning.}
\label{tab:exp_1}
\begin{tabularx}{\columnwidth}{l X c}
\toprule
\textbf{Damage Class} & \textbf{Model} & \textbf{MSE $\pm$ Std} \\
\midrule
\multirow{3}{*}{\textbf{Overall}} 
 & \textbf{\dmv~(Ours)} & \textbf{0.50} $\pm$ \textbf{1.29} \\
 & CleanTransformer & 1.59 $\pm$ 1.55 \\
 & \anycar~(Zero-Shot) & 2.59 $\pm$ 2.44 \\
\midrule
\multirow{6}{*}{\textbf{Fall}} 
 & \textbf{\dmv~(Ours)} & \textbf{1.52} $\pm$ \textbf{2.48} \\
 & CleanTransformer & 3.85 $\pm$ 2.46 \\
 & \anycar~(Zero-Shot) & 5.24 $\pm$ 3.66 \\
 & \anycar~(20s Fine-tuning) & 5.20 $\pm$ 3.37 \\
 & \anycar~(5m Fine-tuning) & 3.15 $\pm$ 3.33 \\
 & \anycar~(10m Fine-tuning) & 2.76 $\pm$ 3.13 \\
\midrule
\multirow{6}{*}{\shortstack[l]{\textbf{Broken}\\\textbf{Axle}}} 
 & \textbf{\dmv~(Ours)} & \textbf{0.26} $\pm$ \textbf{0.80} \\
 & CleanTransformer & 0.97 $\pm$ 0.74 \\
 & \anycar~(Zero-Shot) & 2.02 $\pm$ 1.40 \\
 & \anycar~(20s Fine-tuning) & 2.51 $\pm$ 1.90 \\
 & \anycar~(5m Fine-tuning) & 1.89 $\pm$ 1.67 \\
 & \anycar~(10m Fine-tuning) & 1.73 $\pm$ 1.71 \\
\midrule
\multirow{3}{*}{\shortstack[l]{\textbf{Multiple Tires}\\\textbf{Punctured \&}\\\textbf{Suspensions Broken}}} 
 & \textbf{\dmv~(Ours)} & \textbf{0.32} $\pm$ \textbf{0.13} \\
 & CleanTransformer & 1.57 $\pm$ 0.32 \\
 & \anycar~(Zero-Shot) & 3.39 $\pm$ 2.08 \\
\midrule
\multirow{3}{*}{\shortstack[l]{\textbf{Punctured}\\\textbf{Tire}}} 
 & \textbf{\dmv~(Ours)} & \textbf{0.19} $\pm$ \textbf{0.16} \\
 & CleanTransformer & 0.91 $\pm$ 0.13 \\
 & \anycar~(Zero-Shot) & 1.32 $\pm$ 0.96 \\
\midrule
\multirow{3}{*}{\shortstack[l]{\textbf{Broken}\\\textbf{Suspension}}} 
 & \textbf{\dmv~(Ours)} & \textbf{0.21} $\pm$ \textbf{0.20} \\
 & CleanTransformer & 0.92 $\pm$ 0.15 \\
 & \anycar~(Zero-Shot) & 1.67 $\pm$ 1.57 \\
 \midrule
\multirow{3}{*}{\shortstack[l]{\textbf{No}\\\textbf{Damage}}} 
 & \textbf{\dmv~(Ours)} & \textbf{0.45} $\pm$ \textbf{1.14} \\
 & CleanTransformer & 1.38 $\pm$ 0.80 \\
 & \anycar~(Zero-Shot) & 2.08 $\pm$ 1.61 \\
\bottomrule
\end{tabularx}
\end{table}

Results presented in Table~\ref{tab:exp_1} show \dmv~significantly outperforms all baselines across every damage class.
Most notably, \dmv~achieves superior prediction accuracy in a zero-shot manner, effectively generalizing to unseen damages without requiring any online data collection or weight updates.

In contrast, \anycar~struggles to generalize to the altered dynamics in zero-shot.
While the fine-tuned \anycar~variants show progressive improvement as the duration of online interaction increases from 20 seconds to 10 minutes, they fail to match the precision of \dmv.
Even after 10 minutes of data collection, which involves significant risk when operating a mechanically compromised platform, the fine-tuned models still yield higher prediction errors than our instant, language-conditioned approach.
Furthermore, CleanTransformer performs poorly across all scenarios.
Despite possessing a similar backbone as our method, its inability to account for the structural health state leads to substantial prediction errors.

\subsection{Research Question 2: Architectural Necessity}

To validate the architectural necessity of our 6-DoF state decomposition, we compare \dmv~against a Monolithic Transformer baseline.
We use a Transformer Encoder-Decoder architecture that takes damage embedding as input but processes the robot's state as a single token, obscuring the independent relationships between dimensions.

\begin{figure}[t]
    \centering
    \includegraphics[width=\columnwidth]{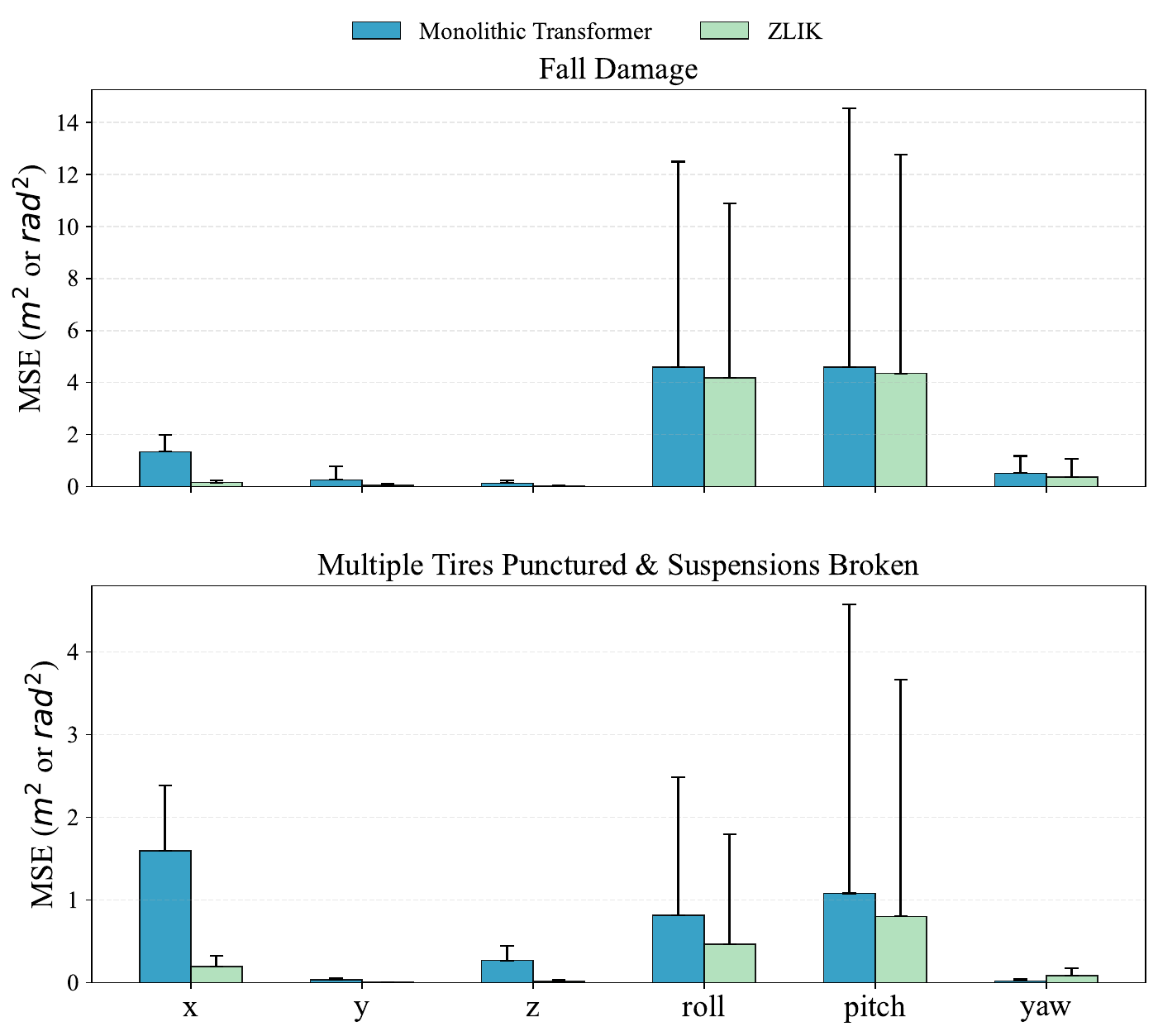}
    \caption{\dmv~outperforms Monolithic Transformer in terms of MSE and standard deviation across state dimensions for Fall and Multiple Tires Punctured \& Suspensions Broken.}
    \label{fig:state_decomposition}
\end{figure}

Table~\ref{tab:ablation_state_decomp} shows the comparison of MSE and its standard deviation across all classes. 
The Monolithic Transformer yields larger prediction error.
This trend is also evident in Fig.~\ref{fig:state_decomposition}, where we compare the errors in the decomposed states in damage classes with multiple mechanical failures (Fall and Multiple Tires Punctured \& Suspensions Broken).
This highlights that without state decomposition, the model cannot effectively disentangle the complex, non-linear disturbances induced by vehicle damage.

\begin{table}[h]
    \centering
    \caption{Impact of State Decomposition on Trajectory Prediction Accuracy for all Damage Classes. The Monolithic Transformer without state decomposition, yields larger error.}
    \label{tab:ablation_state_decomp}
    \begin{tabular}{lcc}
        \toprule
        \textbf{Model Architecture} & \textbf{State Decomposition} & \textbf{MSE $\pm$ Std} \\
        \midrule
        Monolithic Transformer& $\times$ & 0.81 $\pm$ 1.11 \\
        \textbf{\dmv\ (Ours)} & \checkmark & \textbf{0.50} $\pm$ \textbf{1.09} \\
        \bottomrule
    \end{tabular}
\end{table}

\subsection{Research Question 3: Semantic Grounding}

We structure this experiment to evaluate \dmv's ability to successfully align linguistic semantics and physical behaviors.
Specifically, we test the hypothesis that the model uses semantic description to achieve high prediction accuracy.
We choose trials of four distinct damage classes from the 30K unseen test set. 
Trials with matching descriptions are compared against trials where an incorrect description is randomly assigned.

The confusion matrix (Table~\ref{tab:kinodynamics_results}) shows a clear diagonal dominance, indicating that prediction error is minimized when the input language description matches the physical state of the robot.
When the model is provided with an incorrect description for a given damage, the prediction error increases significantly. This validates that the model is effectively using the language-informed semantic embedding $\mathbf{z}_x$ to condition its kinodynamics prediction.

The fourth column of the confusion matrix highlights an interesting observation.
Due to the severity of failures in the Multiple Tires Punctured \& Suspensions Broken class, the maneuverability of such a damaged robot is significantly restricted.
As the vehicle's range of possible motion is reduced, kinodynamic prediction becomes inherently easier, resulting in lower overall errors,
considering the representation of state-action history can provide sufficient prior to bias the future state predictions. 
However, even in classes with low prediction errors, the inclusion of the correct damage embedding provides a measurable positive impact.
This demonstrates that even when the dynamics are constrained by severe damages, semantic grounding allows the model to capture subtle independent nuances in independent DoFs.


\begin{table}[ht]
\centering
\small 
\caption{The confusion matrix evaluates \dmv's ability to ground linguistic semantics in physical behaviors. The diagonal dominance demonstrates that the model successfully utilizes the semantic embedding $\mathbf{z}_x^{t}$ to condition kinodynamics.}
\label{tab:kinodynamics_results}
\begin{tabularx}{\columnwidth}{Xccc|c} 
\toprule
 & \multicolumn{4}{c}{\textbf{Actual Physical State}} \\
\cmidrule(lr){2-5}
\textbf{Input Language} & \textbf{Broken} & \multirow{2}{*}{\textbf{Fall}}  & \textbf{No} & \textbf{MTP} \\
\textbf{Description} & \textbf{Axle} &  & \textbf{Damage} & \textbf{\& SB}  \\
\midrule
\textbf{Broken Axle} & \textbf{0.28} & 0.38 & 0.29 & 0.21 \\
Description & \textbf{($\pm$0.82)} & ($\pm$0.88) & ($\pm$0.61) & ($\pm$0.41) \\
\addlinespace
\textbf{Fall} & 0.34 & \textbf{0.31} & 0.36 & 0.23 \\
Description & ($\pm$0.84) & \textbf{($\pm$0.85)} & ($\pm$0.63) & ($\pm$0.48) \\
\addlinespace
\textbf{No Damage} & 0.29 & 0.39 & \textbf{0.28} & 0.22 \\
Description & ($\pm$0.83) & ($\pm$0.88) & \textbf{($\pm$0.61)} & ($\pm$0.47) \\
\addlinespace
\textbf{MTP \& SB }& 0.29 & 0.37 & 0.29 & \textbf{0.19} \\
Description & ($\pm$0.82) & ($\pm$0.88) & ($\pm$0.60) & \textbf{($\pm$0.39)} \\
\bottomrule
\end{tabularx}
\end{table}

\subsection{Research Question 4: Cross-Embodiment Generalization}

\begin{figure}[t]
    \centering
    \includegraphics[width=\columnwidth]{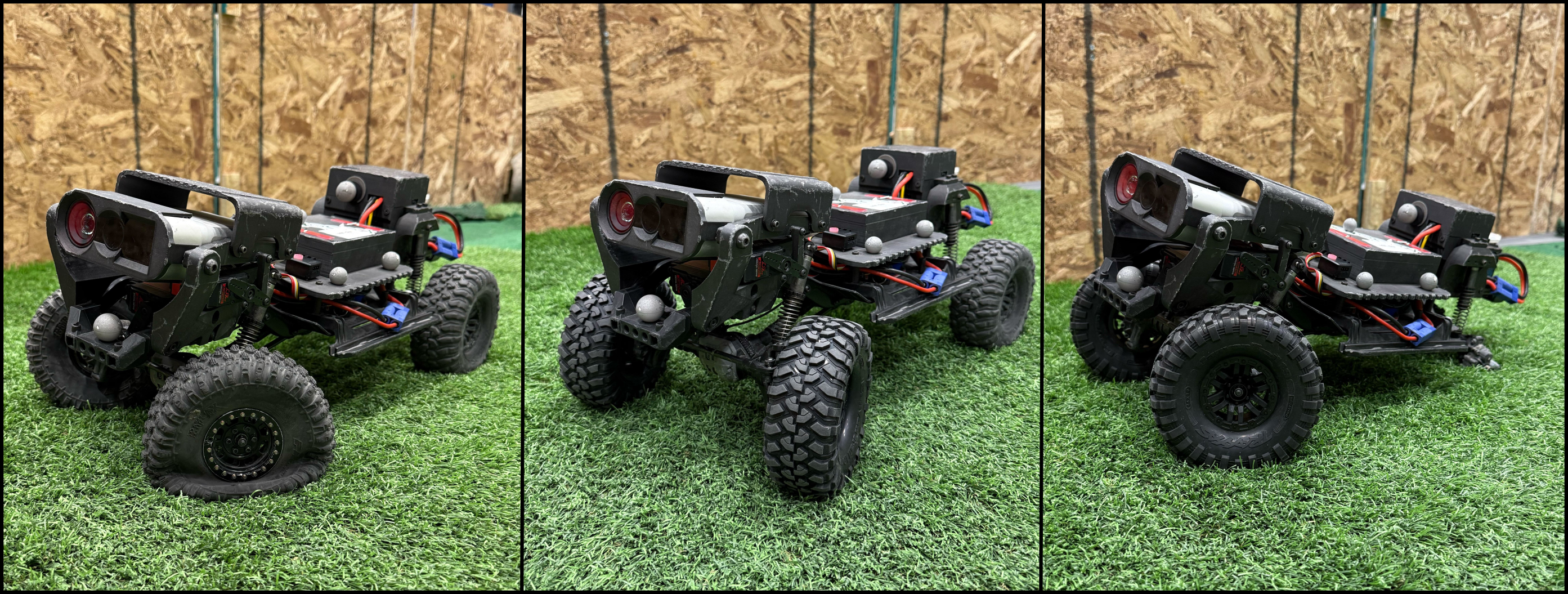}
    \caption{Healthy Physical Robot Platform, V4W (middle), Used in the Cross-Embodiment Generalization Experiment, Compared with the Damaged V4W Variants, Front Left Tire Punctured (left) and Rear Left Wheel Removed (right).} 
    \label{fig:v4w}
\end{figure}

\begin{figure}[t]
    \centering
    \includegraphics[width=\columnwidth]{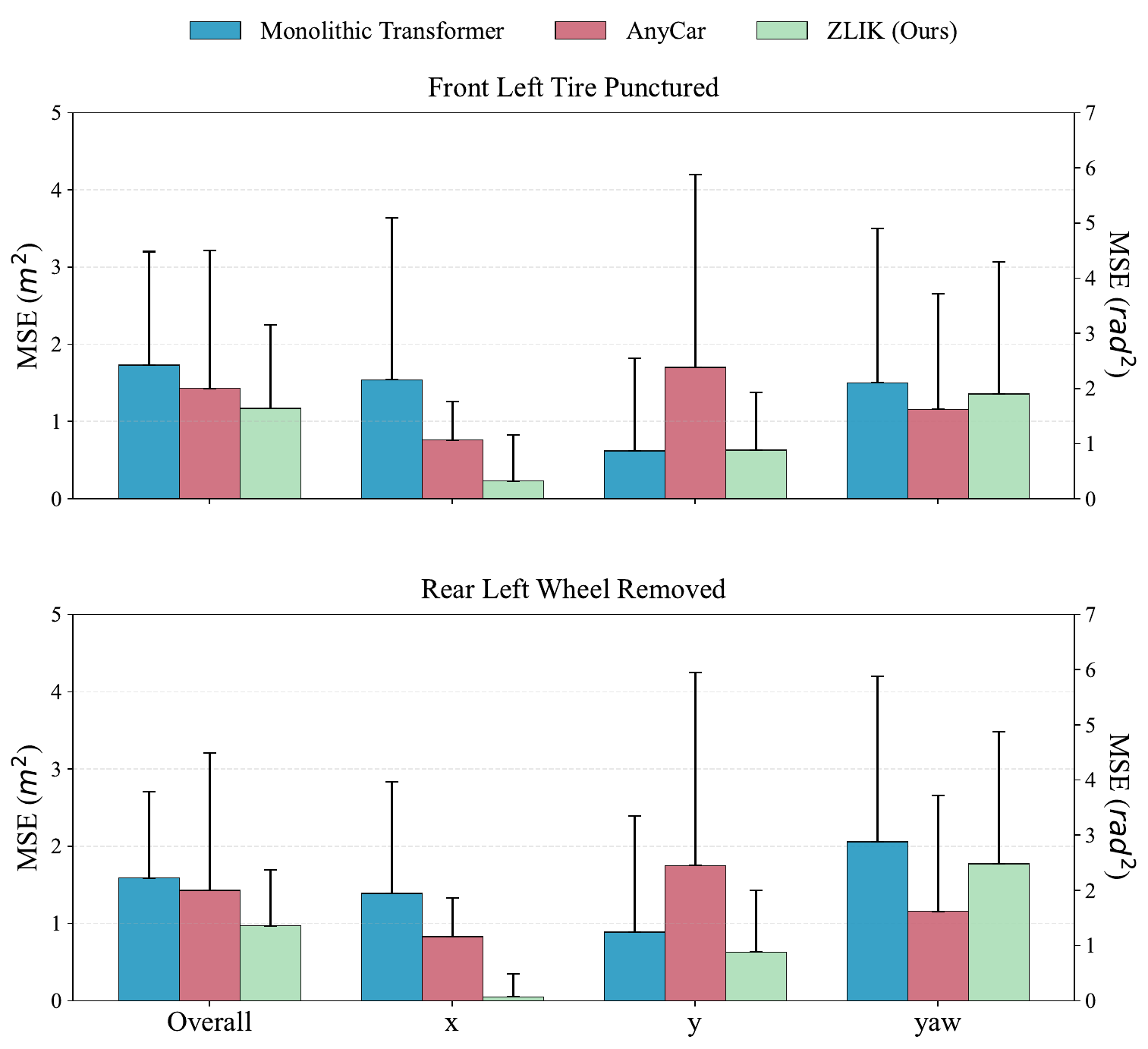}
    \caption{Cross-Embodiment Evaluation on a Physical 1/10$^{\text{th}}$ Scale V4W Robot. \dmv~achieves the best performance in zero-shot on two unseen damage classes in most dimensions.} 
    \label{fig:cross-embodiment}
\end{figure}

This experiment introduces a large domain shift as the test robot differs from the training robot not only in terms of the deployment environment (simulation vs real world), but also  fundamental physical parameters like mass, dimensions, wheelbase, and actuation limits.
To validate \dmv, we test our model on a 1/10$^{\text{th}}$ scale physical V4W platform (Fig.~\ref{fig:v4w})~\cite{datar2024terrain}.
We alter the robot's health state by removing the foam inside the front left tire to simulate a punctured tire and removing the rear left wheel to simulate a detached wheel.
The V4W with altered health states suffers from non-nominal kinodynamic behaviors compared to a healthy V4W platform, exhibiting the problem we aim to address with \dmv.
We compare the prediction error of \dmv~with the CleanTransformer, the Monolithic Transformer, and the \anycar~model, as shown in Fig.~\ref{fig:cross-embodiment}. 

Transitioning from a full-sized simulated vehicle to a 1/10$^{\text{th}}$ scale physical robot inevitably introduces a performance gap.
We observe that the relative error rate increases for all models on the physical platform compared to simulation.
This increase is expected due to unmodeled physical factors, such as sensor noise, communication latency, and friction disparity inherent to the sim-to-real and full-to-1/10$^{\text{th}}$ scale gaps.

Despite the increase in relative error rate, \dmv~demonstrates improvement over the baselines in a zero-shot fashion. This suggests that our semantic embeddings capture the fundamental nature of the damage (e.g., ``loss of a wheel leads to drag on that side") rather than overfitting to the full-size training vehicle in simulation. Even for the unseen 1/10$^{\text{th}}$ scale physical test robot, the semantic grounding provides a prior that is robust enough to outperform the three baselines, effectively bridging the gap across different vehicle embodiments.

\section{CONCLUSIONS} 
\label{sec::conclusions}
We present \dmv, a natural language-informed kinodynamics modeling approach that can adapt to robot structural damage in a zero-shot manner. By grounding linguistic semantics in physical behaviors, \dmv\ precisely predicts kinodynamics when facing a variety of vehicle damages without any online data collection and retraining, even crossing the sim-to-real as well as full-to-1/10$^{\text{th}}$ scale gaps. One potential avenue for future work is to consider terrain geometry and semantics in the kinodynamics model, as such environmental features also play a vital role in determining the changes in kinodynamics when facing structural damages. 
\section*{ACKNOWLEDGEMENT}


This work has taken place in the RobotiXX Laboratory at George Mason University. RobotiXX research is supported by National Science Foundation (NSF, 2350352), Army Research Office (ARO, W911NF2320004, W911NF2420027, W911NF2520011), Air Force Research Laboratory (AFRL), US Air Forces Central (AFCENT), Google DeepMind (GDM), Clearpath Robotics, Raytheon Technologies (RTX), Tangenta, Mason Innovation Exchange (MIX), and Walmart.

\bibliographystyle{plainnat}
\bibliography{mybib}

@article{xiao2021learning,
  title={Learning inverse kinodynamics for accurate high-speed off-road navigation on unstructured terrain},
  author={Xiao, Xuesu and Biswas, Joydeep and Stone, Peter},
  journal={IEEE Robotics and Automation Letters},
  volume={6},
  number={3},
  pages={6054--6060},
  year={2021},
  publisher={IEEE}
}

@article{meng2023terrainnet,
  title={TerrainNet: Visual Modeling of Complex Terrain for High-speed, Off-road Navigation},
  author={Meng, Xiangyun and Hatch, Nathan and Lambert, Alexander and Li, Anqi and Wagener, Nolan and Schmittle, Matthew and Lee, JoonHo and Yuan, Wentao and Chen, Zoey and Deng, Samuel and others},
  journal={arXiv preprint arXiv:2303.15771},
  year={2023}
}

@inproceedings{karnan2022vi,
  title={Vi-ikd: High-speed accurate off-road navigation using learned visual-inertial inverse kinodynamics},
  author={Karnan, Haresh and Sikand, Kavan Singh and Atreya, Pranav and Rabiee, Sadegh and Xiao, Xuesu and Warnell, Garrett and Stone, Peter and Biswas, Joydeep},
  booktitle={2022 IEEE/RSJ International Conference on Intelligent Robots and Systems (IROS)},
  pages={3294--3301},
  year={2022},
  organization={IEEE}
}

@inproceedings{rabiee2019friction,
  title={A friction-based kinematic model for skid-steer wheeled mobile robots},
  author={Rabiee, Sadegh and Biswas, Joydeep},
  booktitle={2019 International Conference on Robotics and Automation (ICRA)},
  pages={8563--8569},
  year={2019},
  organization={IEEE}
}

@article{datar2023learning,
  title={Learning to model and plan for wheeled mobility on vertically challenging terrain},
  author={Datar, Aniket and Pan, Chenhui and Xiao, Xuesu},
  journal={IEEE Robotics and Automation Letters},
  year={2024},
  publisher={IEEE}
}

@INPROCEEDINGS{He2016,
  author={He, Kaiming and Zhang, Xiangyu and Ren, Shaoqing and Sun, Jian},
  booktitle={2016 IEEE Conference on Computer Vision and Pattern Recognition (CVPR)}, 
  title={Deep Residual Learning for Image Recognition}, 
  year={2016},
  pages={770-778},
  doi={10.1109/CVPR.2016.90}
}

@article{maheshwari2023piaug,
  title={PIAug--Physics Informed Augmentation for Learning Vehicle Dynamics for Off-Road Navigation},
  author={Maheshwari, Parv and Wang, Wenshan and Triest, Samuel and Sivaprakasam, Matthew and Aich, Shubhra and Rogers III, John G and Gregory, Jason M and Scherer, Sebastian},
  journal={arXiv preprint arXiv:2311.00815},
  year={2023}
}

@article{han2023model,
  title={Model predictive control for aggressive driving over uneven terrain},
  author={Han, Tyler and Liu, Alex and Li, Anqi and Spitzer, Alex and Shi, Guanya and Boots, Byron},
  journal={arXiv preprint arXiv:2311.12284},
  year={2023}
}

@inproceedings{ning2025dkmgp,
  author       = {Jingyun Ning and
                  Madhur Behl},
  editor       = {Necmiye Ozay and
                  Laura Balzano and
                  Dimitra Panagou and
                  Alessandro Abate},
  title        = {{DKMGP:} {A} Gaussian Process Approach to Multi-Task and Multi-Step
                  Vehicle Dynamics Modeling in Autonomous Racing},
  booktitle    = {7th Annual Learning for Dynamics {\&} Control Conference},
  series       = {Proceedings of Machine Learning Research},
  volume       = {283},
  pages        = {59--71},
  publisher    = {{PMLR}},
  year         = {2025}
}

@article{ostafew2016robust,
author = {Chris J. Ostafew and Angela P. Schoellig and Timothy D. Barfoot},
title ={Robust Constrained Learning-based NMPC enabling reliable mobile robot path tracking},
journal = {The International Journal of Robotics Research},
volume = {35},
number = {13},
pages = {1547-1563},
year = {2016},
doi = {10.1177/0278364916645661},
URL = {https://doi.org/10.1177/0278364916645661},
eprint = {https://doi.org/10.1177/0278364916645661}
}

@inproceedings{gibson2025dynamics,
  title={Dynamics modeling using visual terrain features for high-speed autonomous off-road driving},
  author={Gibson, Jason and Alavilli, Anoushka and Tevere, Erica and Theodorou, Evangelos A and Spieler, Patrick},
  booktitle={2025 IEEE International Conference on Robotics and Automation (ICRA)},
  pages={9809--9815},
  year={2025},
  organization={IEEE}
}

@inproceedings{zhao2024physord,
  title={PhysORD: a neuro-symbolic approach for physics-infused motion prediction in off-road driving},
  author={Zhao, Zhipeng and Li, Bowen and Du, Yi and Fu, Taimeng and Wang, Chen},
  booktitle={2024 IEEE/RSJ International Conference on Intelligent Robots and Systems (IROS)},
  pages={11670--11677},
  year={2024},
  organization={IEEE}
}

@article{cai2025pietra,
  title={Pietra: Physics-informed evidential learning for traversing out-of-distribution terrain},
  author={Cai, Xiaoyi and Queeney, James and Xu, Tong and Datar, Aniket and Pan, Chenhui and Miller, Max and Flather, Ashton and Osteen, Philip R and Roy, Nicholas and Xiao, Xuesu and others},
  journal={IEEE Robotics and Automation Letters},
  year={2025},
  publisher={IEEE}
}

@inproceedings{wang2024pay,
  title={Pay attention to how you drive: Safe and adaptive model-based reinforcement learning for off-road driving},
  author={Wang, Sean J and Zhu, Honghao and Johnson, Aaron M},
  booktitle={2024 IEEE International Conference on Robotics and Automation (ICRA)},
  pages={16954--16960},
  year={2024},
  organization={IEEE}
}

@article{lupu2024magic,
  title={MAGIC VFM-meta-learning adaptation for ground interaction control with visual foundation models},
  author={Lupu, Elena Sorina and Xie, Fengze and Preiss, James Alan and Alindogan, Jedidiah and Anderson, Matthew and Chung, Soon-Jo},
  journal={IEEE Transactions on Robotics},
  volume={41},
  pages={180--199},
  year={2024},
  publisher={IEEE}
}

@article{levy2025meta,
  title={Meta-Learning Online Dynamics Model Adaptation in Off-Road Autonomous Driving},
  author={Levy, Jacob and Gibson, Jason and Vlahov, Bogdan and Tevere, Erica and Theodorou, Evangelos and Fridovich-Keil, David and Spieler, Patrick},
  journal={arXiv preprint arXiv:2504.16923},
  year={2025}
}

@inproceedings{finn2017model,
  title={Model-agnostic meta-learning for fast adaptation of deep networks},
  author={Finn, Chelsea and Abbeel, Pieter and Levine, Sergey},
  booktitle={International conference on machine learning},
  pages={1126--1135},
  year={2017},
  organization={PMLR}
}

@article{nagabandi2018learning,
  title={Learning to adapt in dynamic, real-world environments through meta-reinforcement learning},
  author={Nagabandi, Anusha and Clavera, Ignasi and Liu, Simin and Fearing, Ronald S and Abbeel, Pieter and Levine, Sergey and Finn, Chelsea},
  journal={arXiv preprint arXiv:1803.11347},
  year={2018}
}

@article{grigsby2021long,
  title={Long-range transformers for dynamic spatiotemporal forecasting},
  author={Grigsby, Jake and Wang, Zhe and Nguyen, Nam and Qi, Yanjun},
  journal={arXiv preprint arXiv:2109.12218},
  year={2021}
}

@inproceedings{
zhang2023crossformer,
title={Crossformer: Transformer Utilizing Cross-Dimension Dependency for Multivariate Time Series Forecasting},
author={Yunhao Zhang and Junchi Yan},
booktitle={The Eleventh International Conference on Learning Representations },
year={2023},
url={https://openreview.net/forum?id=vSVLM2j9eie}
}

@article{nagy2023ensemble,
  title={Ensemble gaussian processes for adaptive autonomous driving on multi-friction surfaces},
  author={Nagy, Tom{\'a}{\v{s}} and Amine, Ahmad and Nghiem, Truong X and Rosolia, Ugo and Zang, Zirui and Mangharam, Rahul},
  journal={IFAC-PapersOnLine},
  volume={56},
  number={2},
  pages={494--500},
  year={2023},
  publisher={Elsevier}
}

@inproceedings{yu2023fully,
  author={Yu, Xihang and Teng, Sangli and Chakhachiro, Theodor and Tong, Wenzhe and Li, Tingjun and Lin, Tzu-Yuan and Koehler, Sarah and Ahumada, Manuel and Walls, Jeffrey M. and Ghaffari, Maani},
  booktitle={2023 IEEE/RSJ International Conference on Intelligent Robots and Systems (IROS)}, 
  title={Fully Proprioceptive Slip-Velocity-Aware State Estimation for Mobile Robots via Invariant Kalman Filtering and Disturbance Observer}, 
  year={2023},
  volume={},
  number={},
  pages={8096-8103},
  doi={10.1109/IROS55552.2023.10342519}
}

@article{Pacejka01011992,
author = {Hans B. Pacejka and Egbert Bakker},
title = {THE MAGIC FORMULA TYRE MODEL},
journal = {Vehicle System Dynamics},
volume = {21},
number = {sup001},
pages = {1--18},
year = {1992},
publisher = {Taylor \& Francis},
doi = {10.1080/00423119208969994},
URL = { 
        https://doi.org/10.1080/00423119208969994
},
eprint = { 
        https://doi.org/10.1080/00423119208969994

}
}

@book{rajamani2006vehicle,
  title={Vehicle dynamics and control},
  author={Rajamani, Rajesh},
  year={2006},
  publisher={Springer}
}

@article{williams2018information,
  title={Information-theoretic model predictive control: Theory and applications to autonomous driving},
  author={Williams, Grady and Drews, Paul and Goldfain, Brian and Rehg, James M and Theodorou, Evangelos A},
  journal={IEEE Transactions on Robotics},
  volume={34},
  number={6},
  pages={1603--1622},
  year={2018},
  publisher={IEEE}
}

@inproceedings{nazeri2025verticoder,
  title={Verticoder: Self-supervised kinodynamic representation learning on vertically challenging terrain},
  author={Nazeri, Mohammad and Datar, Aniket and Pokhrel, Anuj and Pan, Chenhui and Warnell, Garrett and Xiao, Xuesu},
  booktitle={2025 IEEE International Conference on Robotics and Automation (ICRA)},
  pages={6536--6543},
  year={2025},
  organization={IEEE}
}

@inproceedings{datar2024terrain,
  title={Terrain-attentive learning for efficient 6-DoF kinodynamic modeling on vertically challenging terrain},
  author={Datar, Aniket and Pan, Chenhui and Nazeri, Mohammad and Pokhrel, Anuj and Xiao, Xuesu},
  booktitle={2024 IEEE/RSJ International Conference on Intelligent Robots and Systems (IROS)},
  pages={5438--5443},
  year={2024},
  organization={IEEE}
}

@inproceedings{datar2024toward,
  title={Toward wheeled mobility on vertically challenging terrain: Platforms, datasets, and algorithms},
  author={Datar, Aniket and Pan, Chenhui and Nazeri, Mohammad and Xiao, Xuesu},
  booktitle={2024 IEEE International Conference on Robotics and Automation (ICRA)},
  pages={16322--16329},
  year={2024},
  organization={IEEE}
}

@article{nazeri2025vertiformer,
  title={VertiFormer: A Data-Efficient Multi-Task Transformer for Off-Road Robot Mobility},
  author={Nazeri, Mohammad and Pokhrel, Anuj and Card, Alexandyr and Datar, Aniket and Warnell, Garrett and Xiao, Xuesu},
  journal={arXiv preprint arXiv:2502.00543},
  year={2025}
}

@article{raissi2019physics,
  title={Physics-informed neural networks: A deep learning framework for solving forward and inverse problems involving nonlinear partial differential equations},
  journal={Journal of Computational physics},
  volume={378},
  pages={686--707},
  year={2019},
  publisher={Elsevier}
}

@article{ovadia2019can,
  title={Can you trust your model's uncertainty? evaluating predictive uncertainty under dataset shift},
  author={Ovadia, Yaniv and Fertig, Emily and Ren, Jie and Nado, Zachary and Sculley, David and Nowozin, Sebastian and Dillon, Joshua and Lakshminarayanan, Balaji and Snoek, Jasper},
  journal={Advances in neural information processing systems},
  volume={32},
  year={2019}
}

@article{vaswani2017attention,
  title={Attention is all you need},
  author={Vaswani, Ashish and Shazeer, Noam and Parmar, Niki and Uszkoreit, Jakob and Jones, Llion and Gomez, Aidan N and Kaiser, {\L}ukasz and Polosukhin, Illia},
  journal={Advances in neural information processing systems},
  volume={30},
  year={2017}
}

@inproceedings{wigness2019rugd,
  author={Wigness, Maggie and Eum, Sungmin and Rogers, John G. and Han, David and Kwon, Heesung},
  booktitle={2019 IEEE/RSJ International Conference on Intelligent Robots and Systems (IROS)}, 
  title={A RUGD Dataset for Autonomous Navigation and Visual Perception in Unstructured Outdoor Environments}, 
  year={2019},
  volume={},
  number={},
  pages={5000-5007},
  doi={10.1109/IROS40897.2019.8968283}
}

@inproceedings{xiao2025anycar,
  title={Anycar to anywhere: Learning universal dynamics model for agile and adaptive mobility},
  author={Xiao, Wenli and Xue, Haoru and Tao, Tony and Kalaria, Dvij and Dolan, John M and Shi, Guanya},
  booktitle={2025 IEEE International Conference on Robotics and Automation (ICRA)},
  pages={8819--8825},
  year={2025},
  organization={IEEE}
}

@software{beamng_tech,
    title = "{B}eam{NG}.tech",
    author = {{BeamNG GmbH}},
    url = {https://www.beamng.tech/},
    version = {0.35.0.0},
    date = {2025-06-15},
}

@article{remonda2024simulation,
  title={A simulation benchmark for autonomous racing with large-scale human data},
  author={Remonda, Adrian and Hansen, Nicklas and Raji, Ayoub and Musiu, Nicola and Bertogna, Marko and Veas, Eduardo and Wang, Xiaolong},
  journal={Advances in Neural Information Processing Systems},
  volume={37},
  pages={102078--102100},
  year={2024}
}

@article{makoviychuk2021isaac,
  title={Isaac gym: High performance gpu-based physics simulation for robot learning},
  author={Makoviychuk, Viktor and Wawrzyniak, Lukasz and Guo, Yunrong and Lu, Michelle and Storey, Kier and Macklin, Miles and Hoeller, David and Rudin, Nikita and Allshire, Arthur and Handa, Ankur and others},
  journal={arXiv preprint arXiv:2108.10470},
  year={2021}
}

@inproceedings{tasora2016chrono,
  title={Chrono: An open source multi-physics dynamics engine},
  author={Tasora, Alessandro and Serban, Radu and Mazhar, Hammad and Pazouki, Arman and Melanz, Daniel and Fleischmann, Jonathan and Taylor, Michael and Sugiyama, Hiroyuki and Negrut, Dan},
  booktitle={High Performance Computing in Science and Engineering: Second International Conference, HPCSE 2015, Sol{\'a}{\v{n}}, Czech Republic, May 25-28, 2015, Revised Selected Papers 2},
  pages={19--49},
  year={2016},
  organization={Springer}
}

@inproceedings{todorov2012mujoco,
  author={Todorov, Emanuel and Erez, Tom and Tassa, Yuval},
  booktitle={2012 IEEE/RSJ International Conference on Intelligent Robots and Systems}, 
  title={MuJoCo: A physics engine for model-based control}, 
  year={2012},
  volume={},
  number={},
  pages={5026-5033},
  doi={10.1109/IROS.2012.6386109}}

@article{pokhrel2024cahsor,
  title={{CAHSOR}: Competence-Aware High-Speed Off-Road Ground Navigation in {SE}(3)},
  author={Pokhrel, Anuj and Nazeri, Mohammad and Datar, Aniket and Xiao, Xuesu},
  journal={IEEE Robotics and Automation Letters},
  year={2024},
  publisher={IEEE}
}

@inproceedings{lotfi2024uncertainty,
  author={Lotfi, Faraz and Virji, Khalil and Faraji, Farnoosh and Berry, Lucas and Holliday, Andrew and Meger, David and Dudek, Gregory},
  booktitle={2024 IEEE International Conference on Robotics and Automation (ICRA)}, 
  title={Uncertainty-aware hybrid paradigm of nonlinear MPC and model-based RL for offroad navigation: Exploration of transformers in the predictive model}, 
  year={2024},
  volume={},
  number={},
  pages={2925-2931},
  doi={10.1109/ICRA57147.2024.10610452}}

@article{ingebrand2025function,
  title={Function encoders: A principled approach to transfer learning in hilbert spaces},
  author={Ingebrand, Tyler and Thorpe, Adam J and Topcu, Ufuk},
  journal={arXiv preprint arXiv:2501.18373},
  year={2025}
}

@article{cully2015robots,
  title={Robots that can adapt like animals},
  author={Cully, Antoine and Clune, Jeff and Tarapore, Danesh and Mouret, Jean-Baptiste},
  journal={Nature},
  volume={521},
  number={7553},
  pages={503--507},
  year={2015},
  publisher={Nature Publishing Group UK London}
}

@inproceedings{rakelly2019efficient,
  title={Efficient off-policy meta-reinforcement learning via probabilistic context variables},
  author={Rakelly, Kate and Zhou, Aurick and Finn, Chelsea and Levine, Sergey and Quillen, Deirdre},
  booktitle={International conference on machine learning},
  pages={5331--5340},
  year={2019},
  organization={PMLR}
}

@article{xie2021segformer,
  title={SegFormer: Simple and efficient design for semantic segmentation with transformers},
  author={Xie, Enze and Wang, Wenhai and Yu, Zhiding and Anandkumar, Anima and Alvarez, Jose M and Luo, Ping},
  journal={Advances in neural information processing systems},
  volume={34},
  pages={12077--12090},
  year={2021}
}

@inproceedings{jung2024v,
  title={V-strong: Visual self-supervised traversability learning for off-road navigation},
  author={Jung, Sanghun and Lee, JoonHo and Meng, Xiangyun and Boots, Byron and Lambert, Alexander},
  booktitle={2024 IEEE International Conference on Robotics and Automation (ICRA)},
  pages={1766--1773},
  year={2024},
  organization={IEEE}
}

@inproceedings{shaban2022semantic,
  title={Semantic terrain classification for off-road autonomous driving},
  author={Shaban, Amirreza and Meng, Xiangyun and Lee, JoonHo and Boots, Byron and Fox, Dieter},
  booktitle={Conference on Robot Learning},
  pages={619--629},
  year={2022},
  organization={PMLR}
}

@inproceedings{lee2023learning-based,
  title={Learning-based Uncertainty-aware Navigation in 3D Off-Road Terrains},
  author={Lee, Hojin and Kwon, Junsung and Kwon, Cheolhyeon},
  booktitle={2023 IEEE International Conference on Robotics and Automation (ICRA)},
  pages={10061--10068},
  year={2023},
  organization={IEEE}
}

@inproceedings{triest2024velociraptor,
  title={Velociraptor: Leveraging visual foundation models for label-free, risk-aware off-road navigation},
  author={Triest, Samuel and Sivaprakasam, Matthew and Aich, Shubhra and Fan, David and Wang, Wenshan and Scherer, Sebastian},
  booktitle={8th Annual Conference on Robot Learning},
  year={2024}
}

@inproceedings{kirillov2023segment,
  title={Segment anything},
  author={Kirillov, Alexander and Mintun, Eric and Ravi, Nikhila and Mao, Hanzi and Rolland, Chloe and Gustafson, Laura and Xiao, Tete and Whitehead, Spencer and Berg, Alexander C and Lo, Wan-Yen and others},
  booktitle={Proceedings of the IEEE/CVF international conference on computer vision},
  pages={4015--4026},
  year={2023}
}

@article{yang2024depth,
  title={Depth anything v2},
  author={Yang, Lihe and Kang, Bingyi and Huang, Zilong and Zhao, Zhen and Xu, Xiaogang and Feng, Jiashi and Zhao, Hengshuang},
  journal={Advances in Neural Information Processing Systems},
  volume={37},
  pages={21875--21911},
  year={2024}
}

@inproceedings{elnoor2025vlm,
  title={VLM-GroNav: Robot Navigation Using Physically Grounded Vision-Language Models in Outdoor Environments},
  author={Elnoor, Mohamed and Weerakoon, Kasun and Seneviratne, Gershom and Xian, Ruiqi and Guan, Tianrui and Jaffar, Mohamed Khalid M and Rajagopal, Vignesh and Manocha, Dinesh},
  booktitle={2025 IEEE International Conference on Robotics and Automation (ICRA)},
  pages={2391--2398},
  year={2025},
  organization={IEEE}
}

@article{min2025advancing,
  title={Advancing Off-Road Autonomous Driving: The Large-Scale ORAD-3D Dataset and Comprehensive Benchmarks},
  author={Min, Chen and Mei, Jilin and Zhai, Heng and Wang, Shuai and Sun, Tong and Kong, Fanjie and Li, Haoyang and Mao, Fangyuan and Liu, Fuyang and Wang, Shuo and others},
  journal={arXiv preprint arXiv:2510.16500},
  year={2025}
}

@inproceedings{shah2023lm,
  title={Lm-nav: Robotic navigation with large pre-trained models of language, vision, and action},
  author={Shah, Dhruv and Osi{\'n}ski, B{\l}a{\.z}ej and Levine, Sergey and others},
  booktitle={Conference on robot learning},
  pages={492--504},
  year={2023},
  organization={PMLR}
}

@article{kawaharazuka2025vision,
  title={Vision-language-action models for robotics: A review towards real-world applications},
  author={Kawaharazuka, Kento and Oh, Jihoon and Yamada, Jun and Posner, Ingmar and Zhu, Yuke},
  journal={IEEE Access},
  year={2025},
  publisher={IEEE}
}

@article{brohan2023rt,
  title={RT-1: Robotics Transformer for Real-World Control at Scale},
  author={Brohan, Anthony and Brown, Noah and Carbajal, Justice and Chebotar, Yevgen and Dabis, Joseph and Finn, Chelsea and Gopalakrishnan, Keerthana and Hausman, Karol and Herzog, Alexander and Hsu, Jasmine and others},
  journal={Robotics: Science and Systems XIX},
  year={2023},
  publisher={Robotics: Science and Systems Foundation}
}

@inproceedings{radford2021learning,
  title={Learning transferable visual models from natural language supervision},
  author={Radford, Alec and Kim, Jong Wook and Hallacy, Chris and Ramesh, Aditya and Goh, Gabriel and Agarwal, Sandhini and Sastry, Girish and Askell, Amanda and Mishkin, Pamela and Clark, Jack and others},
  booktitle={International conference on machine learning},
  pages={8748--8763},
  year={2021},
  organization={PmLR}
}

@article{liu2023visual,
  title={Visual instruction tuning},
  author={Liu, Haotian and Li, Chunyuan and Wu, Qingyang and Lee, Yong Jae},
  journal={Advances in neural information processing systems},
  volume={36},
  pages={34892--34916},
  year={2023}
}

@article{zhang2026vision,
  title={A Vision-Language-Action Model with Visual Prompt for OFF-Road Autonomous Driving},
  author={Zhang, Liangdong and Nie, Yiming and Li, Haoyang and Kong, Fanjie and Zhang, Baobao and Huang, Shunxin and Fu, Kai and Min, Chen and Xiao, Liang},
  journal={arXiv preprint arXiv:2601.03519},
  year={2026}
}

@ARTICLE{11197900,
  author={Hu, Zechen and Xu, Tong and Xiao, Xuesu and Wang, Xuan},
  journal={IEEE Robotics and Automation Letters}, 
  title={CARoL: Context-Aware Adaptation for Robot Learning}, 
  year={2025},
  volume={10},
  number={11},
  pages={12063-12070},
  doi={10.1109/LRA.2025.3619779}}

@article{bardes2021vicreg,
  title={Vicreg: Variance-invariance-covariance regularization for self-supervised learning},
  author={Bardes, Adrien and Ponce, Jean and LeCun, Yann},
  journal={arXiv preprint arXiv:2105.04906},
  year={2021}
}

@article{vera2025embeddinggemma,
  title={Embeddinggemma: Powerful and lightweight text representations},
  author={Vera, Henrique Schechter and Dua, Sahil and Zhang, Biao and Salz, Daniel and Mullins, Ryan and Panyam, Sindhu Raghuram and Smoot, Sara and Naim, Iftekhar and Zou, Joe and Chen, Feiyang and others},
  journal={arXiv preprint arXiv:2509.20354},
  year={2025}
}

\clearpage
\appendix

\subsection{Training Dataset}
\label{app:dataset_details}

We collect the training dataset across six damage classes in the BeamNG.tech simulator. The total dataset comprises 153,387 trajectory data points, distributed as follows:

\begin{table}[h]
\centering
\caption{Training Dataset Trajectory Distribution}
\begin{tabular}{lr}
\hline
\textbf{Damage Class} & \textbf{No. of Trajectories} \\ \hline
No Damage & 17,955 \\
Tire Puncture & 20,007 \\
Broken Suspension & 25,308 \\
Multiple Tires Punctured \& Suspensions Broken & 27,873 \\
Fall & 46,854 \\
Broken Axle & 15,390 \\ \hline
\textbf{Total} & \textbf{153,387} \\ \hline
\end{tabular}
\end{table}

\subsection{Simulation Experiment Dataset}
For \textbf{Research Questions 1 and 2}, we evaluate the model using a different robot configuration to test zero-shot adaptability:
\begin{itemize}
    \item \textbf{Robot Mass:} 1315 kg
    \item \textbf{Drivetrain:} Front-Wheel Drive (FWD)
    \item \textbf{Test Data:} 15,390 trajectory data points with uniform class distribution.
\end{itemize}

For \textbf{Research Question 3}, we evaluate the model in a test set generated on the same robot used for training data generation.
\begin{itemize}
    \item \textbf{Robot Mass:} 1888 kg
    \item \textbf{Drivetrain:} 4-Wheel Drive (4WD)
    \item \textbf{Test Data:} Uniformly sampled 15,390 trajectories to match the size of previous experiments.
\end{itemize}

\subsection{V4W Dataset}
The real-world validation is conducted on the V4W platform at a sampling rate of 20Hz. The data collected for the two damage classes is summarized in Table \ref{tab:real_world_data}.

\begin{table}[h]
\centering
\caption{Real-World Dataset (V4W Platform)}
\label{tab:real_world_data}
\begin{tabular}{lcc}
\hline
\textbf{Damage Class} & \textbf{Trajectories} & \textbf{Approx. Duration (min)} \\ \hline
Front-Left Puncture & 4,705 & 3.92 \\
Rear-Left Removed & 4,355 & 3.63 \\ \hline
\end{tabular}
\end{table}

\end{document}